# Semantic Segmentation for Fully Automated Macrofouling Analysis on Coatings after Field Exposure

**Lutz M. K. Krause[a], Emily Manderfeld[a], Patricia Gnutt[a], Louisa Vogler[a], Ann Wassick[b], Kailey Richard[b], Marco Rudolph[c], Kelli Z. Hunsucker[b], Geoffrey W. Swain[b], Bodo Rosenhahn[c], Axel Rosenhahn[a]\***

**Affiliations**

[a]Analytical Chemistry – Biointerfaces, Ruhr University Bochum, 44780 Bochum, Germany

[b]Center for Corrosion and Biofouling Control, Florida Institute of Technology, Melbourne, Florida 32901, USA

[c]Institute for Information Processing, Leibniz University Hannover, 30167 Hannover, Germany

\* Corresponding author: axel.rosenhahn@rub.de

**Abstract**

Biofouling is a major challenge for sustainable shipping, filter membranes, heat exchangers, and medical devices. The development of fouling-resistant coatings requires the evaluation of their effectiveness. Such an evaluation is usually based on the assessment of fouling progression after different exposure times to the target medium (e.g., salt water). The manual assessment of macrofouling requires expert knowledge about local fouling communities due to high variances in phenotypical appearance, has single-image sampling inaccuracies for certain species, and lacks spatial information. Here we present an approach for automatic image-based macrofouling analysis. We created a dataset with dense labels prepared from field panel images and propose a convolutional network (adapted U-Net) for the semantic segmentation of different macrofouling classes. The establishment of macrofouling localization allows for the generation of a successional model which enables the determination of direct surface attachment and in-depth epibiotic studies.

# 1 Introduction

Biofouling is the process of the undesired accumulation of biological matter and organisms, which occurs whenever an artificial surface and a living system get in contact. It causes an increase in friction, energy consumption, and running maintenance costs and reduces the efficiency and lifespan of machines such as ships, filter membranes, and heat exchangers. Furthermore, it induces infections during medical treatments and medical device failure resulting in high costs every year (Bixler and Bhushan 2012). Thus, biofouling is still one of the main challenges for modern materials exposed to aquatic environments (Flemming 2011).

The large diversity of biofouling species (Holm 2012) is a major difficulty as their sizes cover a particularly large range (Nurioglu et al. 2015) spanning from nanometers for conditioning films (e.g., organic molecules, proteins), over micrometers for microfouling (e.g., bacteria, diatoms, cells), to millimeters and centimeters for human-visible macrofouling (e.g., barnacles, bryozoans, seaweed). As these colonization stages do not follow a linear "successional" model, the prevention of biofilm formation does not always inhibit macrofoulers from colonizing a clean surface (Callow and Callow 2011). Therefore, a precise analysis of the fouling community on all size scales over long periods, usually months or years, is necessary to evaluate the performance of protective coatings. These times series of coated panels, with several panels per coating, per location, and per static or dynamic experiment, produce large amounts of high-resolution image data, which is evaluated by marine biology experts to judge trends. Further, the high diversity (Holm 2012), seasonality (Kerckhof et al. 2010), local conditions (Canning-Clode and Wahl 2009), and intra-species variability affect the occurrence of species and their macroscopic structure, texture, and color making it sometimes difficult to accurately identify the correct fouling type at every pixel. Therefore, the precise assessment of fouling species is a very demanding but crucial task to provide feedback to material developers about the performance of their coatings. In addition to seasonal variations, the ocean warming, water acidification, and changes in gyres have consequences for biofouling communities (Poloczanska and Butler 2009; Dobretsov et al. 2019), their settlement, growth, composition, and production of bioactive molecules. These are indicators for climate change but also demand a fast adaption of current antifouling solutions to these rapidly shifting conditions.

The adsorption of conditioning films on coatings is typically investigated *in vitro* for example by surface plasmon resonance spectroscopy with model biomacromolecules (Pranzetti et al. 2012; Koc et al. 2019) and unavoidable in natural living systems. For microfouling or "slime", there exists numerous automated microscopy based solutions with high accuracies for the detection or classification of foremost diatoms in drinking water samples (Coltelli et al. 2014; Pedraza et al. 2018; Tang et al. 2018; Ruiz-Santaquiteria et al. 2020) and in complex, mixed species environments (Deng et al. 2021) particularly on coatings after field tests (Krause et al. 2020). But for macrofouling, encountering the wide range of size scales and the differences between early and later stages of fouling is necessary. The field of remote sensing offers several solutions for the underwater detection and segmentation of marine organisms growing relatively large on static offshore structures (Gormley et al. 2018) such as corals in benthic communities (Beijbom et al. 2015; King et al. 2018; Alonso et al. 2019; Pavoni et al. 2020) and macroalgae (Balado et al. 2021). In contrast, communities growing under dynamic conditions on primarily moving vessels have a different composition (Bloomfield et al. 2021), smaller sizes, and consist of early growth stages due to periodic cleaning. The fouling of ship hulls is typically investigated by underwater photography (O'Byrne et al. 2020; Peng et al. 2020; Bloomfield et al. 2021; First et al. 2021) to quantify the biofouling coverage and manage the biosecurity risk of invasive species. However, for the analysis of panels with experimental coatings it is common practice to remove them from water for capturing images onshore (Chin et al. 2017). This is fast and inexpensive as scuba divers or remotely operated vehicles (Butler et al. 2009) are not required, but results in a collapsed appearance of the macrofouling species compared to underwater images. Consequently, we limited our work to onshore images of early macrofouling stages where a direct interaction between the artificial surface and the settling organisms can be observed as this is most relevant for material research and the long-term prevention of biofouling. An example of fouled panels and the manual segmentation into fouling classes is shown in Figure 1. Our work aims to automate this process of macrofouling image analysis.

Usually, the percentage cover and composition of biofouling on test panels are either examined by on-site visual assessment by experts following standardized guidelines (ASTM International 2020) or by digital imaging and subsequent random point annotation using Coral Point Count with Excel extensions (Kohler and Gill 2006) (CPCe). For visual assessment, experts directly determine the attached species and visually estimate their coverage on the entire panel. CPCe tries to standardize the latter by sampling typically 50 − 100 uniformly distributed random points over the panel which must be annotated with the fouling species by experts from which percentage coverages per species are calculated. Obviously, the accuracy of both methods relies on the perception and experience of the expert and the knowledge of local fouling communities. Though CPCe is a more quantitative approach, it induces new biases as non-stratified random points cause clustering with large unannotated areas sometimes disregarding entire fouling classes present on panels with diverse fouling communities making sampling error an important consideration (Bloomfield et al. 2021). Both, expert visual assessment and point-based annotation approaches are valuable for the determination of bulk metrics such as percentage cover, but do not provide information to quantify the demographic change or to conduct spatial analysis, which demand semantic segmentation (Pavoni et al. 2020). This offers localizability within time series data which allows for the determination of per-pixel surface attachment of organisms before they become overgrown, which is an exceptionally beneficial metric for material and epibiotic research.

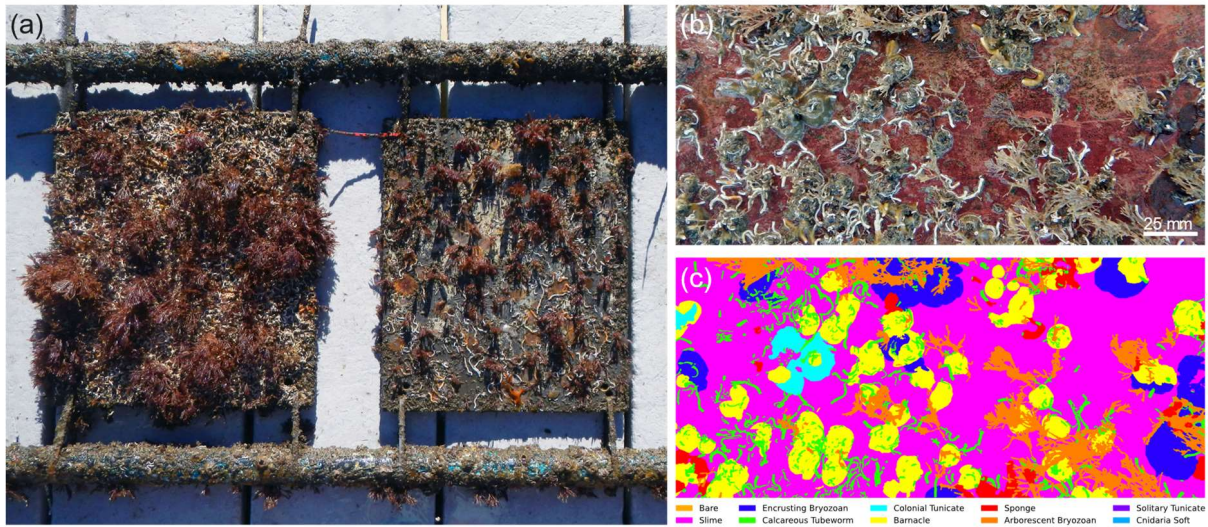

**Figure 1: Biofouling on coated panels and manual semantic segmentation.** Photograph (a) of fouled PVC panels with epoxy coatings mounted on a rack and lifted out of the water after several months of immersion at Port Canaveral in Florida, USA. Cropped photograph (b) of a medium-fouled panel and the indexed color image (c) of its corresponding pixel-wise manual segmentation into fouling classes.

Previous work for the automated analysis of early stage macrofouling images aimed for the classification of biofouling on the image level either for overall determination of the level of fouling (Bloomfield et al. 2021) or single species recognition (Chin et al. 2017). Apart from being mainly unsuitable underwater images, it is generally possible to leverage these image level annotations (Kolesnikov and Lampert 2016) or abundant point annotations (Bearman et al. 2016) for semantic segmentation by weakly supervised learning. Unfortunately, the performance depends on the image level of detail, thus many small (windingly shaped) objects per image reduce the performance of both methods as usually at least one labeled pixel per object is necessary (Alonso et al. 2019). This prerequisite is not given as random point assignment is ambiguous and sparse. Unsupervised clustering into separate fouling classes (First et al. 2021) is hindered by high intra-species variances. Consequently, pixel-accurate semantic segmentation for macrofouling images demands dense segmentation masks and common fully convolutional network (Shelhamer et al. 2017) variants such as U-Net (Ronneberger et al. 2015), SegNet (Badrinarayanan et al. 2017), DeepLab (Chen et al. 2018), or custom adaptions of them. Previous works provide either segmentations for attachment on marine current turbines (Peng et al. 2020) and biofouling on synthetic underwater images (O'Byrne et al. 2018) without further species classification or single species segmentation of tunicates (Galloway et al. 2017) and barnacles (O'Byrne et al. 2020). A more detailed approach is the patch-based segmentation of submerged panel images into seven classes (bare, slime, algae, tunicates, bryozoans, cnidaria, others) using sparse coding and a support vector machine (First et al. 2021). Unfortunately, the training dataset contains 2000 patches of size $9 \times 9$ pixels randomly selected from five heavily fouled underwater panel images with the same coating, which is unsuitable for our task, because of the submerged recording process, the low label detail, the limitation to one coating type, and its specific fouling community.

Hence, the reuse of labeled datasets, weakly supervised learning from random points, and unsupervised clustering are unfeasible demanding the creation of a new dataset with dense labels for the semantic segmentation of onshore macrofouling images. For efficiency reasons, it is common practice to use active learning (Settles 2009) for the selection of informative and representative examples for time-consuming human labeling, which typically results in a performance close to full supervision with fewer labels (Yang et al. 2017; Casanova et al. 2020). Recent approaches make use of data-driven (Konyushkova et al. 2015; Casanova et al. 2020) or hand-crafted heuristics (Vezhnevets et al. 2012; Jain and Grauman 2016; Gorriz et al. 2017; Yang et al. 2017; Sener and Savarese 2018) to control this selection process for semantic segmentation. We employ a previously developed and straightforward pool-based method (Yang et al. 2017) for biomedical image segmentation, which considers the uncertainty of the samples and approximates batch-wise a representative maximum set cover of the unlabeled instances.

Essentially, there exists neither an approach for the detailed semantic segmentation of onshore early stage macrofouling images, which would allow for percentage cover, demographic change, and spatial analysis to assist biofouling researchers to assess coatings and to monitor the impact of climate change on fouling communities, nor an annotated dataset that could be used. Our main contribution can be summarized as follows: (i) we create a dataset with dense labels for semantic segmentation, (ii) we quantify the variability of random point analysis, (iii) we propose a customized U-Net as a benchmark model and demonstrate its performance and generalization, and (iv) we generate layer models for epibiotic analysis and for the determination of direct surface attachment as a novel biofouling metric.

# 2 Material and Methods

## 2.1 Data acquisition

The macrofouling data were acquired through static immersion of panels at three test sites within the Indian River Lagoon system, Florida, USA located in Port Canaveral (28°24′28.76″ N, 80°37′39.11″ W), Melbourne (28°4′36.05″ N, 80°36′1.93″ W) and Grant (27°55′47.32″ N, 80°31′32.15″ W). The test site located in Port Canaveral has the greatest oceanic influence with an average salinity of $34 \pm 1.6$ and average water temperature of $25 \pm 3.8$°C. The Melbourne test site is in the mixing zone of a freshwater creek and the main estuary with an average salinity of $16 \pm 4.6$ and average water temperature of $27 \pm 4.1$°C. The Grant test site is in the main estuary further away from freshwater influences and has an average salinity of $25 \pm 5.1$ and average water temperature of $26 \pm 4.5$°C.

The data were divided into three datasets with varying degrees of sparse annotation and experimental conditions. For all datasets, panels were immersed vertically approximately 0.5 m below the surface and assessed monthly (ASTM International 2020). For Dataset A, 13 x 30 cm panels were coated by manufacturers according to required specifications using antifouling and fouling release control coatings. For Datasets B and C, 25 x 30 cm panels were coated with International® Intergard 264® using a roller and applied to a thickness of 156 μm wet (125 μm dry). Dataset A consisted of 486 images obtained by a Nikon Coolpix AW120 and was annotated with the present macrofouling species at 50 random points by experts (Kohler and Gill 2006; ASTM International 2020). These photographs were divided into 108 images of the Grant and Port Canaveral test sites from 07/2019 and 378 images of the Port Canaveral test site from 01/2020 – 07/2020. For each location and month 54 images of 10 different panel coatings with 2 – 6 replicates were recorded. Dataset B was composed of 16 images obtained by a Nikon Coolpix AW100 and was annotated with percentage coverage by the visual assessment of experts (ASTM International 2020). Eight images were acquired at each of the sites Melbourne and Port Canaveral from 01/2020 – 08/2021. Dataset C was composed of 30 time series images obtained by a Nikon Coolpix AW100 and was annotated with percentage coverage by the visual assessment of experts. At two sites, Melbourne and Port Canaveral, photos of five panels per site of three subsequent months per panel from 10/2019 – 05/2021 were used resulting in individual three-month time series. For a fair verification of our approach, the datasets A and B were fused to dataset A+B and used for dense labeling, training, validation, and active learning. Dataset C was used as an independent dataset for the demonstration of generalization and advanced analysis of assignment accuracy.

## 2.2   Data preprocessing

All images were cropped to obtain single panel pictures. In addition, 12.5 % of the top and bottom and 1 % of the left right image borders were removed to crop the fastening and holes. Images of the small panels were resized to $1472 \times 2752$ px and images of the large panels to $3008 \times 2752$ px. Subsequently the images were enhanced by contrast stretching (25 % cutoff) and slight unsharp masking ($r = 1$, $p = 100\%$). The surface coverage annotation was reduced to ten classes (bare, slime, barnacle, arborescent bryozoan, encrusting bryozoan, colonial tunicate, solitary tunicate, calcareous tubeworms, sponge, cnidaria) with at least 50 random points per class.

The resulting panel images of dataset A+B were sliced to tiles of 384 × 384 px size with 64 px overlap. This is a suitable size to capture sufficient contextual information of large macrofouling organisms such as barnacles or solitary tunicates, but small enough to ensure a good distribution of rarely occurring macrofouling classes. The moderate labeling effort per tile allowed an iterative buildup of the learning dataset while monitoring that also rarely occurring classes are well represented. The slicing generated a pool of over 16,000 unlabeled image tiles, from which samples for annotation were selected.

## 2.3    Annotation and training methodology

For annotation, we decided to label only the very top macrofouling organisms and not buried species directly attached to the surface. For example, if a tubeworm is covered by a transparent yellowish sponge, those pixels are annotated as sponge. This approach is at a first glance in contrast to visual assessment or random point annotation by experts, who usually focus on the species in direct contact with the surface. For the semantic segmentation we are forced to classify only the visible image content, any estimation of underlying layers requires the presented analysis of fouling progression.

Initially, we randomly selected 398 image tiles from the unlabeled pool and asked experts to select further 602 tiles with underrepresented classes, which were mainly barnacles, bare, cnidaria, sponges, and tunicates, for dense labeling. From this labeled set, we generated additional 149 synthetically labeled samples covering the overlap areas of four neighboring tiles. The resulting set of 1,149 tiles was representatively split into 862 tiles for training and 287 for validation. For a balanced split, we used an evolutionary algorithm (Blickle 2000) with tournament selection ($k = 3$) and elitism to minimize the Kullback-Leibler divergence between the training and validation set class distribution. To increase the diversity and representativeness, we expanded the training set by active learning with 400 image tiles following a previously developed method for biomedical images(Yang et al. 2017) with $K = 64$ and $k = 16$. However, we estimate the uncertainty of an image tile by the entropy of its predicted logits instead of the variance among bootstrapped models. In our case, this avoids repeatedly training of several models since the mean entropy of a tile shows a proper correlation with the loss on the validation set.

Image tiles of the dataset A+B were downsampled by a factor of 2 without perceptional changes to decrease the training time and memory consumption and to reduce annotation inaccuracies on object contours. While active learning decreased class imbalances by a preferred selection of rare and consequently uncertain classes, we also utilized oversampling of training set tiles containing minority classes. We calculated the class weights $w_c$ from their normalized probability $p_c$ in the training set and lower bounded the weights to one:

$$w_c = \frac{\max\limits_c p_c}{p_c} \tag{1}$$

The number of additional samples $n_i$ per image tile $i$ was calculated from its normalized class probability $p_{c,i}$ regarding that a tile with an average weight should not be oversampled. We empirically set $\alpha = 2$ which allows for an enhanced control over the process:

$$n_i = \max\left(\left\lceil \sum_c p_{c,i} \cdot w_c - \frac{1}{N} \cdot \sum_i \sum_c p_{c,i} \cdot w_c - \alpha \right\rceil, 0\right) \tag{2}$$

During training we employed data augmentations (flipping, rotation, gray scaling, contrast changes, hue adjustments, elastic transformation, class dropout, coarse dropout, and contrast limited adaptive histogram equalization (Pizer et al. 1990)) to increase the model robustness. We found the optimal strength of augmentations by RandAugment (Cubuk et al. 2020). Learning rate decay and early stopping were also used.

## 2.4   Model

The final model architecture is illustrated in Supplementary Figure 5 and is based on U-Net (Ronneberger et al. 2015) which yielded excellent performance on various biomedical datasets (Moen et al. 2019; Isensee et al. 2021) along with proper hyperparameter tuning. We enriched the U-Net with a pretrained EfficientNet B2 encoder (Tan and Le 2019), residual links in the decoder (He et al. 2016; Jegou et al. 2017), self-attention layers on the channel dimension (Hu et al. 2018) to learn cross-channel correlations, and a rebalanced number of decoder filters per expansion stage. Contrary to the original U-Net, we continuously used bilinear upsampling instead of transposed convolution to avoid checkerboard patterns (Odena et al. 2016), batch normalization (Ioffe and Szegedy 2015), and ReLU6 activation (Krizhevsky and Hinton 2010; Sandler et al. 2018). Hyperparameters were tuned by grid search to achieve the performance reported.

## 2.5 Loss function

The semantic segmentation loss is a combination of the dice loss (Drozdzal et al. 2016; Isensee et al. 2021) and the cross-entropy loss which were calculated separately for each sample in a batch. It is of the form

$$\mathcal{L} = \frac{1}{2} \cdot (\mathcal{L}_{dice} + \mathcal{L}_{CE}) \tag{3}$$

where $\mathcal{L}_{CE}$ is the categorical cross-entropy loss and the dice loss $\mathcal{L}_{dice}$ is of the form

$$\mathcal{L}_{dice} = 1 - \frac{2}{|C|} \sum_{c \in C} \frac{\sum_{i \in I} y_i^c \hat{y}_i^c}{\sum_{i \in I} y_i^c + \hat{y}_i^c} \tag{4}$$

where is $y_i^c$ the one-hot encoded label and $\hat{y}_i^c$ is the softmax prediction of the network for each pixel $i \in I$ of a training sample and each class $c \in C$ for the set of all present classes $C$.

## 2.6 Implementation details

Training was implemented in TensorFlow. The utilized optimizer was Adam (Kingma and Ba 2014) with its default parameters, and a batch size of 16. We used two phases for training, whereby in the first phase only the decoder was trained and in the subsequent phase also higher encoder layers were fine tuned. The initial learning rate of the first phase was $1 \times 10^{-3}$ and $1 \times 10^{-4}$ of the second phase. Whenever the validation loss did not improve by more than $1 \times 10^{-4}$ within the last 30 epochs, the learning rate was decreased by a factor 5 with a lower bound of $1 \times 10^{-6}$. Each training phase was limited to 300 epochs, but early stopping was used to terminate the training if the validation loss did not improve within the last 30 epochs. Mixed precision training until convergence took about 250 epochs in total using a single Nvidia RTX 2070 8GB GPU, an AMD 3700X CPU, and 32 GB of memory.

## 2.7 Random point sampling

For the statistical analysis of the random point annotation approach, we used completely annotated panel images from the dataset A+B containing all considered classes. We simulated the process by uniform sampling of 50 random points per panel image and repeated this procedure 1000 times per image for statistically reliable results. All probabilities and errors were calculated a-posteriori meaning that class-specific results for an image were only considered if the class was present in this image. The class distribution from random points or the complete image was given by their normalized pixel-wise occurrence $p_c$. The resulting mean absolute error (MAE), the mean absolute percentage error (MAPE), and the left-out probability for a class $c$ are of the form:

$$\text{MAE}(c) = \frac{1}{|I^c||N|} \sum_{i \in I^c} \sum_{n \in N} \left| p_{c,i} - \hat{p}_{c,i,n} \right| \tag{5}$$

$$\text{MAPE}(c) = \frac{1}{|I^c||N|} \sum_{i \in I^c} \sum_{n \in N} \frac{\left| p_{c,i} - \hat{p}_{c,i,n} \right|}{p_{c,i}} \tag{6}$$

$$\text{LOP}(c) = \frac{1}{|I^c||N|} \sum_{i \in I^c} \sum_{n \in N \ \&\& \ \hat{p}_{c,i,n} > 0} 1 \tag{7}$$

Where $I^c$ is the set of images containing class $c$, $|N|$ is the number of repetitions per image, $p_{c,i}$ is the relative probability of class $c$ in image $i \in I$, and $\hat{p}_{c,i,n}$ is the relative class probability for the $n$-th sampled set of random points on image $i$.

## 2.8 Latent space visualization

For the visualization of the latent space spanned by the encoder of our network, we extracted the feature maps of every image tile from the highest encoder layer and performed average pooling to obtain an embedding representing the semantic information in the entire tile. We transformed the resulting distribution of all image tiles by channel-wise normalization to mean $\mu = 0$ and standard deviation $\sigma = 1$ and subsequently performed principal component analysis (PCA) to reduce the dimensionality to 50 preserving most of the variance. Then, t-SNE with a perplexity of 50, PCA initialization, a learning rate of 200, and 2000 iterations was used to reduce the PCA components to two dimensions.

# 3 Results

## 3.1 Dataset Preparation

For the demonstration of our approach, we created a new dataset for the semantic segmentation of onshore early stage macrofouling panel images that were acquired monthly at three sites in the Indian River Lagoon system, Florida, USA. The dataset provides dense segmentation masks for ten classes (bare, slime, barnacle, arborescent bryozoan, encrusting bryozoan, colonial tunicate, solitary tunicate, calcareous tubeworms, sponge, cnidaria) containing the dominant macrofouling species. In detail, dataset A+B consists of 502 panel images with percentage coverage annotation, from which we extracted square tiles for dense labeling selected randomly or by experts to emphasize underrepresented species (1,149) and by uncertainty and diversity aware active learning (400). The obtained tile dataset was representatively split into 1,262 tiles for training containing all samples selected by active learning (Yang et al. 2017) and 287 tiles for validation. We further established an independent dataset C consisting of 30 panel images of three-month time series with percentage coverage annotation for the demonstration of generalization and advanced analysis.

## 3.2 Random Point Annotation

For the determination of bulk biofouling metrics such as percentage coverage estimation, random point sampling is a valuable and fast tool to analyze large time series with several replicates to discover trends. As any method that relies on discrete sampling, rare classes face the risk of large potential errors. Using the labeled data, we investigated the accuracy of this method for single panel images by imitating the sampling and assessment process with 50 random points. Supplementary Table 1 shows the results of our statistical study. While the mean absolute error (MAE) between the obtained and the real percentage coverage was usually small (< 5 %), the relative percentage coverage error of species commonly occupying only small areas was elevated as quantified by the mean absolute percentage error (MAPE). Fine stretched or branched organisms like tubeworms or arborescent bryozoans suffered from a misestimation nearly as high as their surface coverage. Clustered or patch-like occurrences in only some image parts as regularly seen for encrusting bryozoans or sponges similarly caused a varying coverage estimation. The left-out probability (LOP) describes how frequently a class was completely disregarded by random points if present in an image. For classes that typically covered the whole panel (bare) or large areas (barnacle, cnidaria, solitary tunicates) the LOP was very low, but up to 50 % for rare and irregularly distributed classes (tubeworms, arborescent bryozoans). These species experience a systematic underestimation from single image analysis. Both problems are also exemplified in Supplementary Figure 1, where in the first trial sponges were entirely missed out and in the second trial tubeworms were severely overestimated if a limited number of 50 random points was used.

**Table 1: Performance of adapted U-Net model architectures.** Results for image tiles from validation set. Best performing configurations for a metric are highlighted.

| Configuration | Accuracy | IoU | F1 | Precision | Recall |
|---|---|---|---|---|---|
| U-Net (baseline) | 0.959 | 0.614 | 0.746 | 0.758 | 0.739 |
| + Pretrained EfficientNet encoder | 0.974 | 0.735 | 0.841 | 0.842 | **0.841** |
| + Residual decoder links | **0.976** | 0.747 | **0.849** | **0.864** | 0.836 |
| + Channel attention & rebalanced decoder filters | **0.976** | **0.749** | **0.849** | 0.861 | 0.840 |

## 3.3   Automated Semantic Segmentation

As an initial solution for the semantic segmentation of macrofouling panel images, we propose a tuned U-Net enhanced with some recent technologies such as a pretrained EfficientNet (Tan and Le 2019) encoder, residual decoder links, and channel attention layers. We also fine-tuned the hyperparameters by simple grid search. Though there are further potential improvements, we intend to use this network as a baseline for the benchmark of this complex dataset. The results of our modifications are shown in Table 1 using the original U-Net as a baseline. The application of transfer learning using a pretrained encoder is common practice for small datasets and gave the greatest improvement of 20 % in intersection of union (IoU) over the baseline. Additional residual links, channel attention, and hyperparameter tuning resulted in minor increases in IoU but allowed for a better awareness of demanding cases and underrepresented classes. Supplementary Figure 2 perceptually shows the performance at random examples from the validation set.

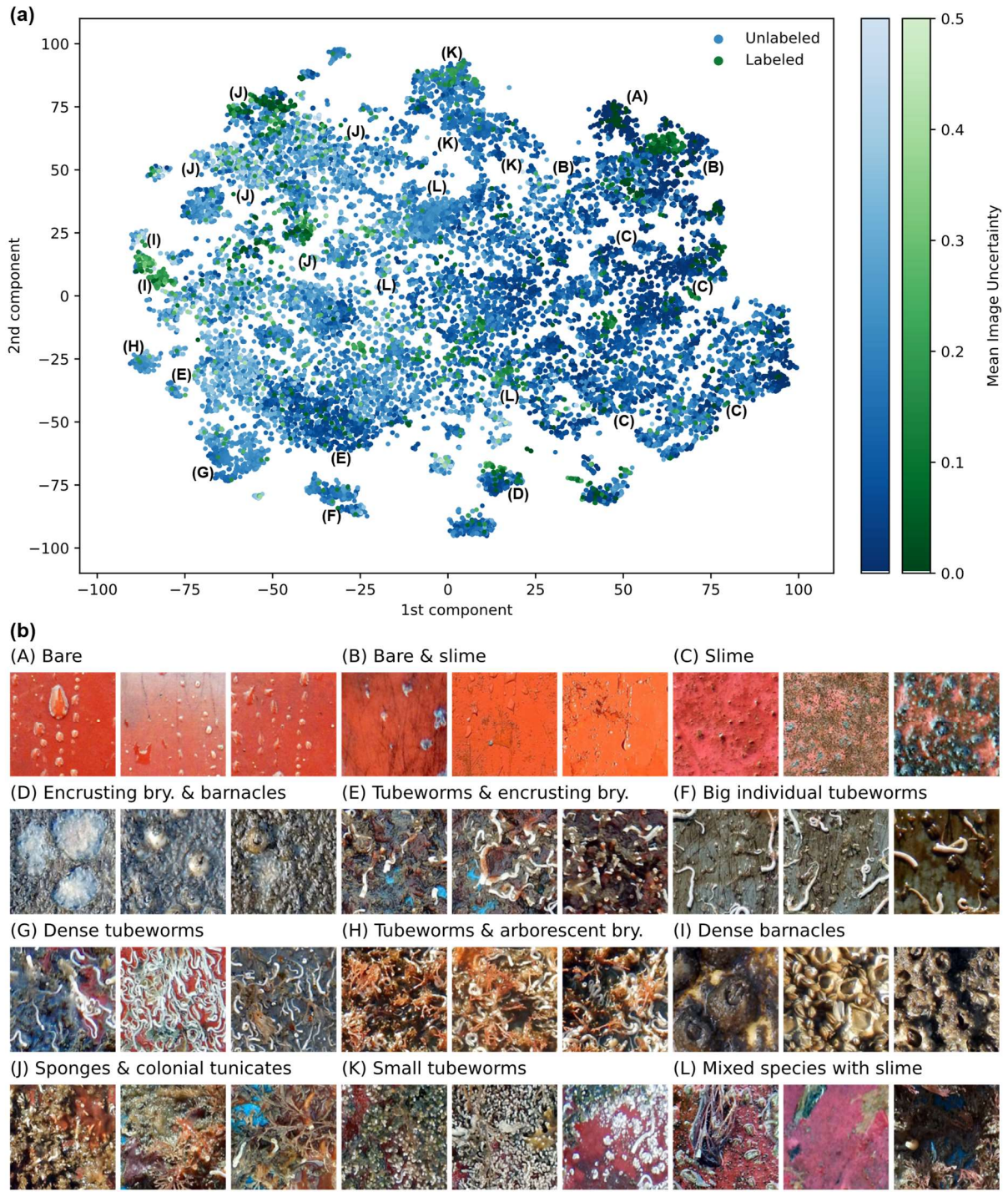

**(a)**

**(b)**

(A) Bare  (B) Bare & slime  (C) Slime

(D) Encrusting bry. & barnacles  (E) Tubeworms & encrusting bry.  (F) Big individual tubeworms

(G) Dense tubeworms  (H) Tubeworms & arborescent bry.  (I) Dense barnacles

(J) Sponges & colonial tunicates  (K) Small tubeworms  (L) Mixed species with slime

**Figure 2: Visualization of the semantic latent space.** Two-dimensional t-SNE visualization (a) of the high-level features extracted by the encoder path of the U-Net and the mean uncertainty per image tile. Random examples (b) of clusters indicated in (a) with similar macrofouling classes but high perceptual variation.

The class-wise performance on the evaluated metrics is shown in Supplementary Table 2. Classes that consist of branched (arborescent bryozoan, cnidaria) or elongated structures (tubeworms) or typically have seamless transition to slime (barnacles) have a decreased IoU of 50 – 60 %, because prediction faults at borders are heavily penalized by this metric. The confusion matrix illustrated in Supplementary

Figure 3 shows a pairwise correlation between slime and arborescent bryozoan, barnacle, cnidaria, colonial tunicate, and tubeworm indicating a slight bias towards slime for human annotators.

The semantic representation of images through a network is decisive for generalization and its interpretation. After random and expert-based tile selection, active learning considered the mean information entropy of predicted logits as image uncertainty and allowed for annotation of ambiguous situations and a smooth sampling throughout the latent space as shown in Figure 2a. Although clusters with a high visual variance led to elevated uncertainty, the network learned a robust understanding of such macrofouling species. An inspection of the largest clusters as shown in Figure 2b revealed a grouping of phenotypes even with high visual variations. While separate growth stages of the same macrofouling species are not necessarily close to each other (e.g., tubeworms in (F) – (H) and (K)), there is usually a smooth transition between co-occurring species (e.g., bare to slime (A) – (C), sponges and colonial tunicates (J)) and different settlement densities of the same species (e.g., for tubeworms low (F), medium (E), (H), and high (G)). For a detailed insight, Supplementary Figure 4 illustrates the spatial distribution of classes categorized by their share of a labeled image tile which emphasized the above findings.

For an elaborated comparison to human experts, we evaluated the prediction of our model to the sparse percentage coverage annotation of time series from dataset A+B. An example of a 7-month time series is shown in Figure 3a where the fast demographic change caused by settlement, growth, dominance, and predatory behavior becomes visible. Figure 3b shows the generalization of our approach among various coatings and replicates. Though human and the network predicted similar trends, the U-Net suffered from continuous underestimation of tubeworms for large surface coverages (coatings F, P) caused by slime overgrowth. Likewise, slime on primarily bare panels (coating H) was underestimated, because of the immense perceptual similarity between slime and wet coating parts or water droplets in the images. Sporadically occurring inequalities in the percentage coverage of sponge and encrusting bryozoan across several replicates stem from challenging seasonal phenotypes and growth stages insufficiently captured by our training set. The remaining trend distinctions were inside the margin of error.

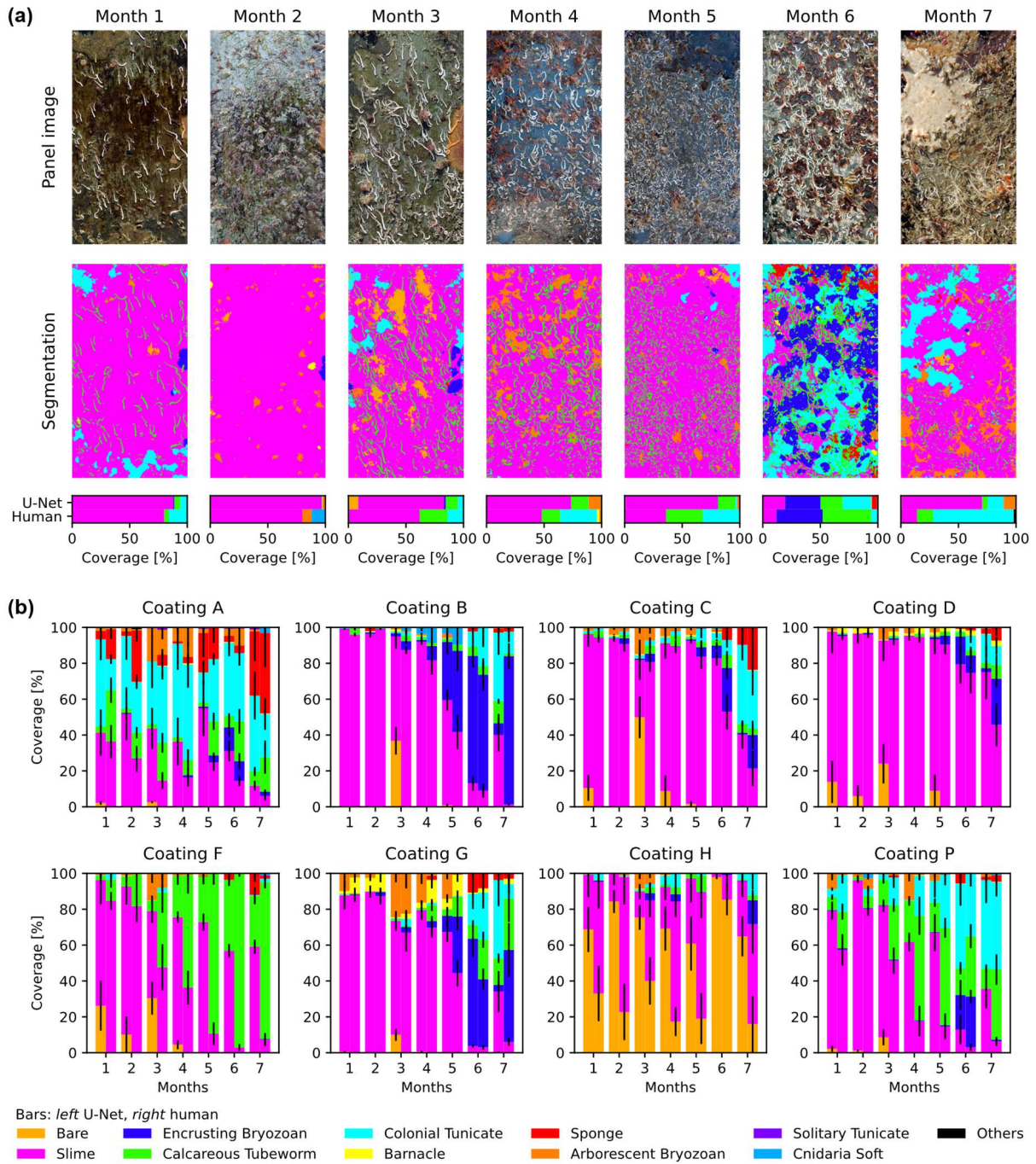

**Figure 3: Comparison of automated and manual analysis.** Seven-month time series (a) of low to moderate fouled panel images with coating P from dataset A+B after immersion at Port Canaveral test site, their segmentation, and the surface coverage acquired by automated and manual estimation. Surface coverage of seven-month time series (b) of fouled panels with different coatings from dataset A+B obtained by U-Net (left bars) and human experts (right bars) after immersion at the Port Canaveral test site. Reported uncertainties refer to the standard error among the $n = 6$ replicates per coating. Coatings with fewer replicates or shorter time series are not shown.

We further evaluated an independent dataset C of the Melbourne test site to show generalization of our approach for the Indian River Lagoon system. The results in Figure 4 show similar trends for human and U-Net surface coverage predictions but with the same underestimation tendencies of tubeworms and bare as described above. An advanced difficulty in dataset C were novel phenotypes of established species like encrusting bryozoan (e.g., Melbourne 2, 3) and cnidaria (e.g., Melbourne 5) which exhibited a completely different morphology and texture. These were misclassified as slime by the U-Net probably caused by the large variance of the slime class and the slight annotation bias towards slime. Unknown classes such as oysters (Melbourne 4) were assigned to their perceptually closest substitutes (i.e., barnacles) and could not be identified by design. In general, this comparison showed the robustness and generalization ability of our approach for known species and similar phenotypes under different experimental conditions.

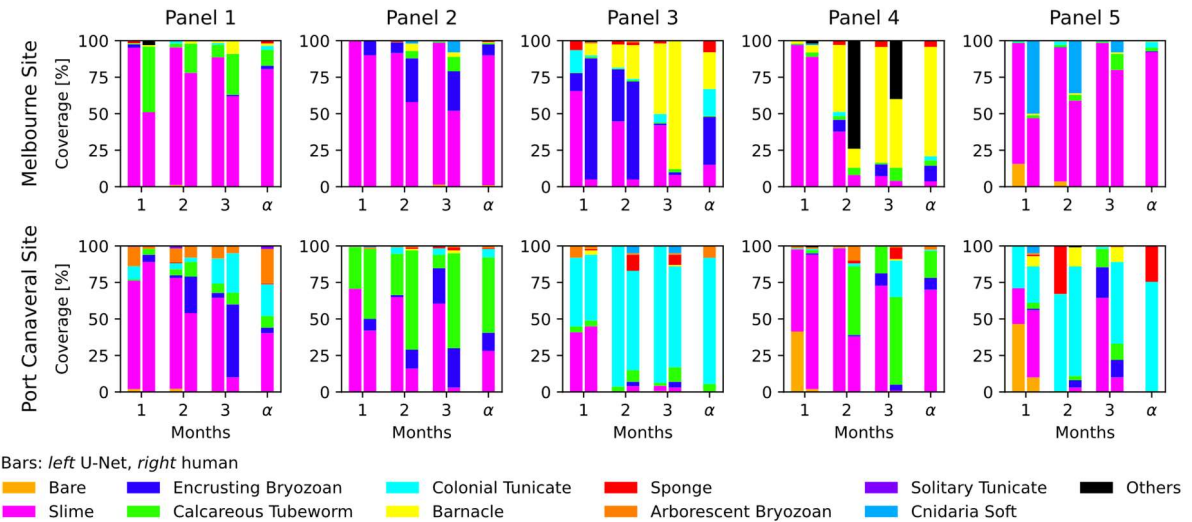

**Figure 4: Comparison of manual and automated analysis.** Panel-wise comparison of the percentage coverage on three-month time series from dataset C analyzed automatically by U-Net (left bars) and by visual assessment of human experts (right bars). The percentage coverage of direct surface attachment (α) was derived from the segmentation masks of each panel-wise time series.

## 3.4   Novel Applications

The opportunity of a precise localization of macrofouling organisms within inferred segmentation masks enables novel applications that are impossible with annotations based on arbitrarily determined and limited sampling points. The images series processed with our model are a time-resolved observation of the species in the top layer on a panel. From this series, we merged information into a layer model (Figure 5c) for each panel to create a spatial representation of fouling progression over time. The layer model was used to compute exclusively the bottom layer (Figure 5a, Figure 4) assuming that the classes slime and bare were successfully dominated by overgrowing species. It enabled the determination of the direct surface attachment which is the percentage coverage of organisms interacting with the developed coating. In the shown example (Figure 5a) species in direct contact with the surface are predominantly tubeworms (52 %), slime (28 %), and encrusting bryozoans (12 %). This is a novel biofouling metric with considerable relevance for marine ecologists and material researchers.

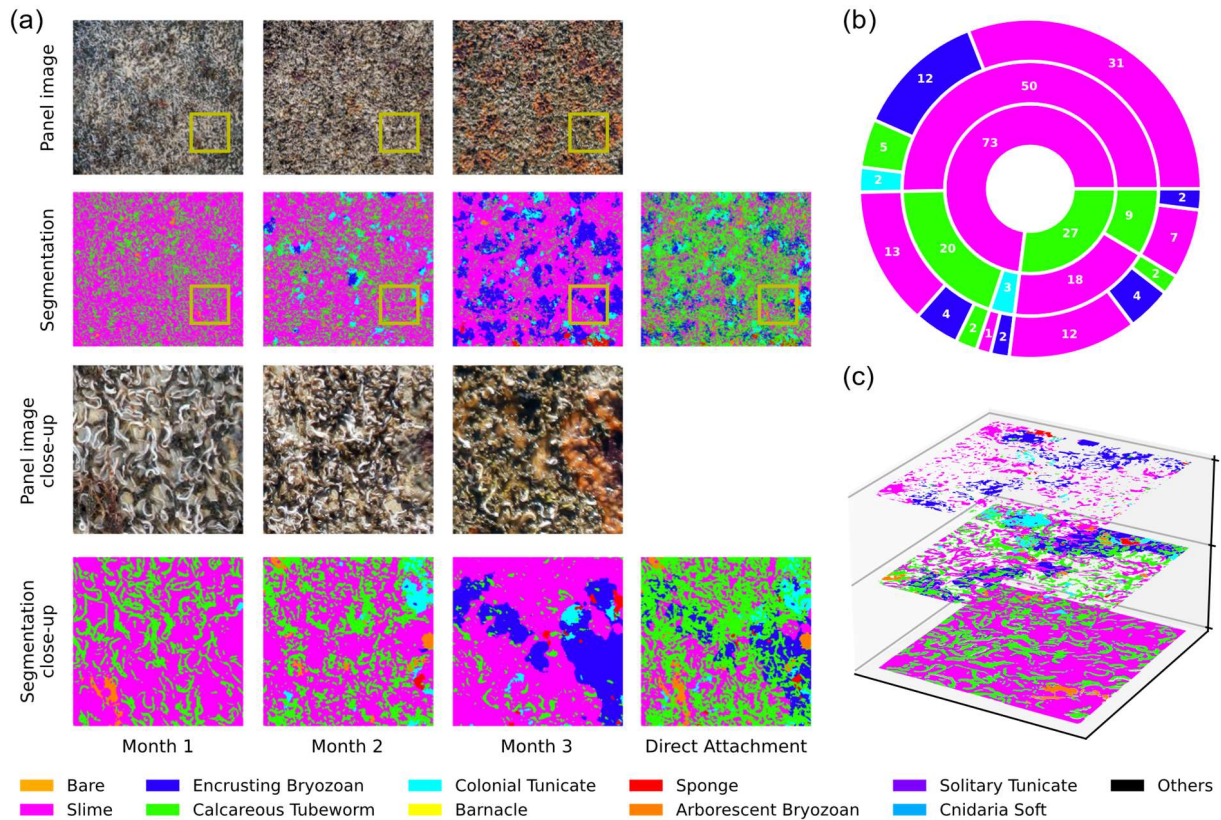

**Figure 5: Determination of macrofoulers in direct coating contact from image time series during fouling progression.** Three-month time series (a) of a moderately-fouled coated panel from dataset C after immersion at the Melbourne test site (first row), its image-wise segmentation (second row), and the generated direct surface attachment (right). Third and fourth row show close-ups of the yellow squares. Succession diagram (b) of the full panel for epibiotic analysis with timeline from inside out. Classes with less than 1 % coverage were removed for better visibility. Three-dimensional projection of the layer model (c) of the close-up exposing successional growth.

The layer model further allowed for an in-depth epibiotic study as illustrated in Figure 5b. While one month after immersion the surface was colonized by slime (73 %) and tubeworms (27 %), proceeding immersion led to an overgrowth. After two months, areas previously covered by slime predominantly remained slime and only a small fraction (27 %) was overgrown by tubeworms. Areas initially covered by tubeworms were majorly (68 %) overgrown by slime. After the third month, additional species occurred. Encrusting bryozoans colonized preferably slime dominated areas (67 %) but also tubeworms (26 %). Tubeworms in turn were primarily overgrown by slime (69 %) and encrusting bryozoans (26 %).

The observation of such relationships and the generation of a successional diagram is exceptionally challenging with common analysis methods and provides valuable epibiotic insight that can be exploited in future studies.

## 4 Discussion

Automated macrofouling analysis on onshore panel images for the quantification of the demographic change or for spatial analysis demands pixel-accurate semantic segmentation (Pavoni et al. 2020), since sparse annotations provide neither the necessary locality nor the prerequisites for weakly supervised learning (Alonso et al. 2019). Thus, we introduced the first dataset for this purpose satisfying the required precision of dense labels and covering the fouling community on different protective coatings. We ensured the annotation of representatives from dataset A+B by smoothly sampling over the entire latent space whilst minimizing uncertainty using active learning as illustrated in Figure 2a. However, as the dataset is currently limited to macrofouling organisms found in Florida and their corresponding phenotypes, it causes a domain bias and should be expanded with data from other sites and fouling classes. The obtained dataset is also of interest for other tasks in naval research as it contains very early and later fouling stages and a high variance in morphological structures, colors, and textures characteristic for marine organisms.

Using the manually segmented images, it was possible to quantify potential inherent sampling inaccuracies in random point assessment for rarely occurring classes such as arborescent and encrusting bryozoans, tubeworms, and sponges. The developed segmentation model can be easily expanded to other test sites and macrofouling species for automated panel evaluation. Our observations on sampling accuracy agree with previous studies that found the sampling inaccuracy of random point annotation for individual images of vessels is not negligible although this method is valuable for percentage cover estimation when multiple replicates are used (Pavoni et al. 2020; Bloomfield et al. 2021). Furthermore, manual analysis methods require expert knowledge about local fouling communities, but it is a time consuming and tiring process (Gormley et al. 2018).

In this work we proposed a method for the semantic segmentation of onshore macrofouling images enhancing previous single species classification (Chin et al. 2017) and overall fouling determination (Bloomfield et al. 2021) approaches. We applied a U-Net with several improvements and coarse hyperparameter optimization, as the straightforward design focus attention on the approach itself and the presented applications to inspire future research in this novel direction. Therefore, we propose our model as a benchmark on this dataset and encourage its improvement and utilization for transfer learning, neighborhood analysis, and fine-grained classification of segmented macrofouling organisms.

Compared to a similar study for underwater coating inspections (First et al. 2021) our approach covered a greater variety of fouling organisms and coatings on onshore images. The mean accuracy of 58 % for seven classes was outperformed by our model with 97.6 % for ten similar classes, but with a different recording process and higher label detail. While our method is currently limited to a subset of the fouling organisms occurring in Florida, it shows similar trends as manual analysis and generalizes over various coating types, phenotypes, and test sites. The robustness observed on dataset C enables the expansion to further macrofouling classes and other sites in the future.

The established possibility of macrofouling localization within time series data allowed for a spatial and demographic analysis which is a huge advantage over manual analysis and reveals new insights for material research, environmental monitoring, and epibiotic studies. The derived layer model and the more specialized direct surface attachment analysis demonstrate the new analytical possibilities. Furthermore, the level of detail of the generated layer model could be controlled by the total immersion duration and the time between consecutive photographs of the panels.

## 5 Conclusions

Semantic segmentation of macrofouling images on test panels is a demanding task requiring large amounts of data and detailed annotations. This work presents a powerful tool for the automated achievement of this task. We hope our approach finds application for the detailed analysis of fouling communities on coatings, their shifts due to ocean acidification, the detection of rare invasive species, environmental monitoring, and hydrodynamic predictions of fouled vessels.

## Data availability

The created semantic segmentation dataset analyzed in this study will be available at https://biointerfaces.ruhr-uni-bochum.de/deep-learning-research/ for non-commercial purposes only.

## Code availability

The code that supports the findings of this study will be available at https://github.com/luuzk/foulingseg.

## Acknowledgements

We thank Labelbox for making their online annotation platform available to us. The work was funded by ONR N00014-20-12244 (Ruhr University Bochum) and ONR N00014-20-1-2243 (Florida Institute of Technology). The authors are grateful to the Deutsche Forschungsgemeinschaft (DFG) for funding in GRK2376/331085229.

## Author contributions

L.K. implemented the study, conducted the experiments, and directed the dataset preparation. E.M., P.G., L.V., A.W., and K.R. acquired and annotated the dataset. All authors contributed to the design of the study, interpreting the results, and writing the manuscript.

## Competing interests

The authors declare no competing interests.

## 6   References

Alonso I, Yuval M, Eyal G, Treibitz T, Murillo AC. 2019. CoralSeg: Learning coral segmentation from sparse annotations. J Field Robotics. 36:1456–1477. doi:10.1002/rob.21915.

ASTM International. 2020. ASTM D6990-20: Standard Practice for Evaluating Biofouling Resistance and Physical Performance of Marine Coating Systems. West Conshohocken, PA: ASTM International  (D6990-20). 2020; [updated 2020]. https://doi.org/10.1520/D6990-20.

Badrinarayanan V, Kendall A, Cipolla R. 2017. SegNet: A Deep Convolutional Encoder-Decoder
Architecture for Image Segmentation. IEEE Trans Pattern Anal Mach Intell. 39:2481–2495. Epub
2017 Jan 2. eng. doi:10.1109/TPAMI.2016.2644615.

Balado J, Olabarria C, Martínez-Sánchez J, Rodríguez-Pérez JR, Pedro A. 2021. Semantic
segmentation of major macroalgae in coastal environments using high-resolution ground imagery
and deep learning. International Journal of Remote Sensing. 42:1785–1800.
doi:10.1080/01431161.2020.1842543.

Bearman A, Russakovsky O, Ferrari V, Fei-Fei L. 2016. What's the Point: Semantic Segmentation
with Point Supervision. In: Leibe B, Matas J, Sebe N, Welling M, editors. Computer Vision –
ECCV 2016. Vol. 9911. Cham: Springer International Publishing; p. 549–565 (Lecture Notes in
Computer Science).

Beijbom O, Edmunds PJ, Roelfsema C, Smith J, Kline DI, Neal BP, Dunlap MJ, Moriarty V, Fan T-Y,
Tan C-J, et al. 2015. Towards Automated Annotation of Benthic Survey Images: Variability of
Human Experts and Operational Modes of Automation. PLoS One. 10:e0130312. Epub 2015 Jul 8.
eng. doi:10.1371/journal.pone.0130312.

Bixler GD, Bhushan B. 2012. Biofouling: lessons from nature. Philos Trans A Math Phys Eng Sci.
370:2381–2417. eng. doi:10.1098/rsta.2011.0502.

Blickle T. 2000. Tournament selection. Evolutionary computation. 1:181–186.

Bloomfield NJ, Wei S, A Woodham B, Wilkinson P, Robinson AP. 2021. Automating the assessment
of biofouling in images using expert agreement as a gold standard. Sci Rep. 11:2739. Epub 2021
Feb 2. eng. doi:10.1038/s41598-021-81011-2.

Butler AJ, Canning-Clode J, Coutts ADM, Cowie PR, Dobretsov S, Drr S, Faimali M, Lewis JA, Page
HM, Pratten J, et al. 2009. Techniques for the Quantification of Biofouling. In: Drr S, Thomason
JC, editors. Biofouling. Oxford, UK: Wiley-Blackwell; p. 319–332.

Callow JA, Callow ME. 2011. Trends in the development of environmentally friendly fouling-resistant
marine coatings. Nat Commun. 2:244. eng. doi:10.1038/ncomms1251.

Canning-Clode J, Wahl M. 2009. Patterns of Fouling on a Global Scale. In: Drr S, Thomason JC, editors. Biofouling. Oxford, UK: Wiley-Blackwell; p. 73–86.

Casanova A, Pinheiro PO, Rostamzadeh N, Pal CJ. 2020. Reinforced active learning for image segmentation. [place unknown]: [publisher unknown]. http://arxiv.org/pdf/2002.06583v1.

Chen L-C, Zhu Y, Papandreou G, Schroff F, Adam H. 2018. Encoder-Decoder with Atrous Separable Convolution for Semantic Image Segmentation. In: Ferrari V, Hebert M, Sminchisescu C, Weiss Y, editors. Computer Vision – ECCV 2018. Vol. 11211. Cham: Springer International Publishing; p. 833–851 (Lecture Notes in Computer Science).

Chin CS, Si J, Clare AS, Ma M. 2017. Intelligent Image Recognition System for Marine Fouling Using Softmax Transfer Learning and Deep Convolutional Neural Networks. Complexity. 2017:1–9. doi:10.1155/2017/5730419.

Coltelli P, Barsanti L, Evangelista V, Frassanito AM, Gualtieri P. 2014. Water monitoring: automated and real time identification and classification of algae using digital microscopy. Environ Sci Process Impacts. 16:2656–2665. eng. doi:10.1039/c4em00451e.

Cubuk ED, Zoph B, Shlens J, Le QV. 2020. Randaugment: Practical automated data augmentation with a reduced search space. In: 2020 IEEE/CVF Conference on Computer Vision and Pattern Recognition Workshops (CVPRW). 2020 IEEE/CVF Conference on Computer Vision and Pattern Recognition Workshops (CVPRW); 14.06.2020 - 19.06.2020; Seattle, WA, USA. [place unknown]: IEEE; p. 3008–3017.

Deng J, Wei H, He D, Gu G, Kang X, Liang H, Liu C, Wu P, Zhong Y, Xu S, et al. 2021. A coarse to fine framework for recognizing and locating multiple diatoms with highly complex backgrounds in forensic investigation. Multimed Tools Appl. doi:10.1007/s11042-021-11169-4.

Dobretsov S, Coutinho R, Rittschof D, Salta M, Ragazzola F, Hellio C. 2019. The oceans are changing: impact of ocean warming and acidification on biofouling communities. Biofouling. 35:585–595. Epub 2019 Jul 8. eng. doi:10.1080/08927014.2019.1624727.

Drozdzal M, Vorontsov E, Chartrand G, Kadoury S, Pal C. 2016. The Importance of Skip Connections in Biomedical Image Segmentation. In: Carneiro G, Mateus D, Peter L, Bradley A, Tavares JMRS, Belagiannis V, Papa JP, Nascimento JC, Loog M, Lu Z, et al., editors. Deep Learning and Data Labeling for Medical Applications. Vol. 10008. Cham: Springer International Publishing; p. 179–187 (Lecture Notes in Computer Science).

First M, Riley S, Islam KA, Hill V, Li J, Zimmerman R, Drake L. 2021. Rapid quantification of biofouling with an inexpensive, underwater camera and image analysis. MBI. 12:599–617. doi:10.3391/mbi.2021.12.3.06.

Flemming H-C. 2011. Microbial Biofouling: Unsolved Problems, Insufficient Approaches, and Possible Solutions. In: Flemming H-C, Wingender J, Szewzyk U, editors. Biofilm Highlights. Vol. 5. Berlin, Heidelberg: Springer Berlin Heidelberg; p. 81–109 (Springer Series on Biofilms).

Galloway A, Taylor GW, Ramsay A, Moussa M. 2017. The Ciona17 Dataset for Semantic Segmentation of Invasive Species in a Marine Aquaculture Environment. In: 2017 14th Conference on Computer and Robot Vision (CRV). 2017 14th Conference on Computer and Robot Vision (CRV); 16.05.2017 - 19.05.2017; Edmonton, AB. [place unknown]: IEEE; p. 361–366.

Gormley K, McLellan F, McCabe C, Hinton C, Ferris J, Kline D, Scott B. 2018. Automated Image Analysis of Offshore Infrastructure Marine Biofouling. JMSE. 6:2. doi:10.3390/jmse6010002.

Gorriz M, Carlier A, Faure E, Giró-i-Nieto X. 2017. Cost-Effective Active Learning for Melanoma Segmentation. CoRR. abs/1711.09168.

He K, Zhang X, Ren S, Sun J. 2016. Identity Mappings in Deep Residual Networks. [place unknown]: [publisher unknown]. http://arxiv.org/pdf/1603.05027v3.

Holm ER. 2012. Barnacles and biofouling. Integr Comp Biol. 52:348–355. Epub 2012 Apr 15. eng. doi:10.1093/icb/ics042.

Hu J, Shen L, Sun G. 2018. Squeeze-and-Excitation Networks. In: 2018 IEEE/CVF Conference on Computer Vision and Pattern Recognition. 2018 IEEE/CVF Conference on Computer Vision and Pattern Recognition (CVPR); 18.06.2018 - 23.06.2018; Salt Lake City, UT. [place unknown]: IEEE; p. 7132–7141.

Ioffe S, Szegedy C. 2015. Batch Normalization: Accelerating Deep Network Training by Reducing Internal Covariate Shift. In: Bach F, Blei D, editors. Proceedings of the 32nd International Conference on Machine Learning. Vol. 37. Lille, France: PMLR; p. 448–456 (Proceedings of Machine Learning Research). https://proceedings.mlr.press/v37/ioffe15.html.

Isensee F, Jaeger PF, Kohl SAA, Petersen J, Maier-Hein KH. 2021. nnU-Net: a self-configuring method for deep learning-based biomedical image segmentation. Nat Methods. 18:203–211. Epub 2020 Dec 7. eng. doi:10.1038/s41592-020-01008-z.

Jain SD, Grauman K. 2016. Active Image Segmentation Propagation. In: 2016 IEEE Conference on Computer Vision and Pattern Recognition (CVPR). 2016 IEEE Conference on Computer Vision and Pattern Recognition (CVPR); 27.06.2016 - 30.06.2016; Las Vegas, NV, USA. [place unknown]: IEEE; p. 2864–2873.

Jegou S, Drozdzal M, Vazquez D, Romero A, Bengio Y. 2017. The One Hundred Layers Tiramisu: Fully Convolutional DenseNets for Semantic Segmentation. In: 2017 IEEE Conference on Computer Vision and Pattern Recognition Workshops (CVPRW). 2017 IEEE Conference on Computer Vision and Pattern Recognition Workshops (CVPRW); 21.07.2017 - 26.07.2017; Honolulu, HI, USA. [place unknown]: IEEE; p. 1175–1183.

Kerckhof F, Rumes B, Norro A, Jacques TG, Degraer S. 2010. Seasonal variation and vertical zonation of the marine biofouling on a concrete offshore windmill foundation on the Thornton Bank (southern North Sea). Offshore wind farms in the Belgian Part of the North Sea: Early environmental impact assessment and spatio-temporal variability. Royal Belgian Institute of Natural Sciences, Management Unit of the North Sea Mathematical Models, Marine ecosystem management unit:53–68.

King A, Bhandarkar SM, Hopkinson BM. 2018. A Comparison of Deep Learning Methods for Semantic Segmentation of Coral Reef Survey Images. In: 2018 IEEE/CVF Conference on Computer Vision and Pattern Recognition Workshops (CVPRW). 2018 IEEE/CVF Conference on Computer Vision and Pattern Recognition Workshops (CVPRW); 18.06.2018 - 22.06.2018; Salt Lake City, UT. [place unknown]: IEEE; p. 1475–14758.

Kingma DP, Ba J. 2014. Adam: A Method for Stochastic Optimization. [place unknown]: [publisher unknown]. http://arxiv.org/pdf/1412.6980v9.

Koc J, Schönemann E, Amuthalingam A, Clarke J, Finlay JA, Clare AS, Laschewsky A, Rosenhahn A. 2019. Low-Fouling Thin Hydrogel Coatings Made of Photo-Cross-Linked Polyzwitterions. Langmuir. 35:1552–1562. Epub 2018 Nov 9. eng. doi:10.1021/acs.langmuir.8b02799.

Kohler KE, Gill SM. 2006. Coral Point Count with Excel extensions (CPCe): A Visual Basic program for the determination of coral and substrate coverage using random point count methodology. Computers & Geosciences. 32:1259–1269. doi:10.1016/j.cageo.2005.11.009.

Kolesnikov A, Lampert CH. 2016. Seed, Expand and Constrain: Three Principles for Weakly-Supervised Image Segmentation. In: Leibe B, Matas J, Sebe N, Welling M, editors. Computer Vision – ECCV 2016. Vol. 9908. Cham: Springer International Publishing; p. 695–711 (Lecture Notes in Computer Science).

Konyushkova K, Sznitman R, Fua P. 2015. Introducing Geometry in Active Learning for Image Segmentation. In: 2015 IEEE International Conference on Computer Vision (ICCV). 2015 IEEE International Conference on Computer Vision (ICCV); 07.12.2015 - 13.12.2015; Santiago, Chile. [place unknown]: IEEE; p. 2974–2982.

Krause LMK, Koc J, Rosenhahn B, Rosenhahn A. 2020. Fully Convolutional Neural Network for Detection and Counting of Diatoms on Coatings after Short-Term Field Exposure. Environ Sci Technol. 54:10022–10030. Epub 2020 Jul 28. eng. doi:10.1021/acs.est.0c01982.

Krizhevsky A, Hinton G. 2010. Convolutional deep belief networks on cifar-10. Unpublished manuscript. 40:1–9.

Moen E, Bannon D, Kudo T, Graf W, Covert M, van Valen D. 2019. Deep learning for cellular image analysis. Nat Methods. 16:1233–1246. Epub 2019 May 27. eng. doi:10.1038/s41592-019-0403-1.

Nurioglu AG, Esteves ACC, With G de. 2015. Non-toxic, non-biocide-release antifouling coatings based on molecular structure design for marine applications. J Mater Chem B. 3:6547–6570. Epub 2015 Jul 8. eng. doi:10.1039/C5TB00232J.

O'Byrne M, Ghosh B, Schoefs F, Pakrashi V. 2020. Applications of Virtual Data in Subsea Inspections. JMSE. 8:328. doi:10.3390/jmse8050328.

O'Byrne M, Pakrashi V, Schoefs F, Ghosh aB. 2018. Semantic Segmentation of Underwater Imagery Using Deep Networks Trained on Synthetic Imagery. JMSE. 6:93. doi:10.3390/jmse6030093.

Odena A, Dumoulin V, Olah C. 2016. Deconvolution and Checkerboard Artifacts. Distill. 1. doi:10.23915/distill.00003.

Pavoni G, Corsini M, Callieri M, Fiameni G, Edwards C, Cignoni P. 2020. On Improving the Training of Models for the Semantic Segmentation of Benthic Communities from Orthographic Imagery. Remote Sensing. 12:3106. doi:10.3390/rs12183106.

Pedraza A, Bueno G, Deniz O, Ruiz-Santaquiteria J, Sanchez C, Blanco S, Borrego-Ramos M, Olenici A, Cristobal G, editors. 2018. Lights and pitfalls of convolutional neural networks for diatom identification. Proc.SPIE. [place unknown]: [publisher unknown] (vol. 10679).

Peng H, Wang T, Pandey S, Chen L, Zhou F. 2020. An Attachment Recognition Method Based on Image Generation and Semantic Segmentation for Marine Current Turbines. In: IECON 2020 The 46th Annual Conference of the IEEE Industrial Electronics Society. IECON 2020 - 46th Annual Conference of the IEEE Industrial Electronics Society; 18.10.2020 - 21.10.2020; Singapore, Singapore. [place unknown]: IEEE; p. 2819–2824.

Pizer SM, Johnston RE, Ericksen JP, Yankaskas BC, Muller KE. 1990. Contrast-limited adaptive histogram equalization: speed and effectiveness. In: [1990] Proceedings of the First Conference on Visualization in Biomedical Computing. [1990] First Conference on Visualization in Biomedical Computing; 22-25 May 1990; Atlanta, GA, USA. [place unknown]: IEEE Comput. Soc. Press; p. 337–345.

Poloczanska ES, Butler AJ. 2009. Biofouling and Climate Change. In: Drr S, Thomason JC, editors. Biofouling. Oxford, UK: Wiley-Blackwell; p. 333–347.

Pranzetti A, Salaün S, Mieszkin S, Callow ME, Callow JA, Preece JA, Mendes PM. 2012. Model Organic Surfaces to Probe Marine Bacterial Adhesion Kinetics by Surface Plasmon Resonance. Adv. Funct. Mater. 22:3672–3681. doi:10.1002/adfm.201103067.

Ronneberger O, Fischer P, Brox T. 2015. U-Net: Convolutional Networks for Biomedical Image Segmentation. In: Navab N, Hornegger J, Wells WM, Frangi AF, editors. Medical Image Computing and Computer-Assisted Intervention – MICCAI 2015. Vol. 9351. Cham: Springer International Publishing; p. 234–241 (Lecture Notes in Computer Science).

Ruiz-Santaquiteria J, Bueno G, Deniz O, Vallez N, Cristobal G. 2020. Semantic versus instance segmentation in microscopic algae detection. Engineering Applications of Artificial Intelligence. 87:103271. doi:10.1016/j.engappai.2019.103271.

Sandler M, Howard A, Zhu M, Zhmoginov A, Chen L-C. 2018. MobileNetV2: Inverted Residuals and Linear Bottlenecks. In: 2018 IEEE/CVF Conference on Computer Vision and Pattern Recognition. 2018 IEEE/CVF Conference on Computer Vision and Pattern Recognition (CVPR); 18.06.2018 - 23.06.2018; Salt Lake City, UT. [place unknown]: IEEE; p. 4510–4520.

Sener O, Savarese S. 2018. Active Learning for Convolutional Neural Networks: A Core-Set Approach. In: International Conference on Learning Representations. [place unknown]: [publisher unknown]. https://openreview.net/forum?id=H1aIuk-RW.

Settles B. 2009. Active Learning Literature Survey. http://digital.library.wisc.edu/1793/60660.

Shelhamer E, Long J, Darrell T. 2017. Fully Convolutional Networks for Semantic Segmentation. IEEE Trans Pattern Anal Mach Intell. 39:640–651. Epub 2016 May 24. eng. doi:10.1109/tpami.2016.2572683.

Tan M, Le Q. 2019. EfficientNet: Rethinking Model Scaling for Convolutional Neural Networks. In: Chaudhuri K, Salakhutdinov R, editors. Proceedings of the 36th International Conference on Machine Learning. Vol. 97. [place unknown]: PMLR; p. 6105–6114 (Proceedings of Machine Learning Research). https://proceedings.mlr.press/v97/tan19a.html.

Tang N, Zhou F, Gu Z, Zheng H, Yu Z, Zheng B. 2018. Unsupervised pixel-wise classification for Chaetoceros image segmentation. Neurocomputing. 318:261–270. doi:10.1016/j.neucom.2018.08.064.

Vezhnevets A, Buhmann JM, Ferrari V. 2012. Active learning for semantic segmentation with expected change. In: 2012 IEEE Conference on Computer Vision and Pattern Recognition. 2012 IEEE Conference on Computer Vision and Pattern Recognition (CVPR); 16.06.2012 - 21.06.2012; Providence, RI. [place unknown]: IEEE; p. 3162–3169.

Yang L, Zhang Y, Chen J, Zhang S, Chen DZ. 2017. Suggestive Annotation: A Deep Active Learning Framework for Biomedical Image Segmentation. In: Descoteaux M, Maier-Hein L, Franz A, Jannin P, Collins DL, Duchesne S, editors. Medical Image Computing and Computer Assisted Intervention − MICCAI 2017. Vol. 10435. Cham: Springer International Publishing; p. 399–407 (Lecture Notes in Computer Science).

# Supplementary Information

**Supplementary Table 1: Error of random point annotation.** The class-wise random point annotation error was quantified by the mean absolute error (MAE), the mean absolute percentage error (MAPE), and the left-out probability (LOP). Metrics were obtained by uniform sampling of $n = 50$ random points with segmentation masks from manual annotation and $N = 1000$ repetitions per image. All values are given as percentage. Reported uncertainties refer to the standard error.

| Class | Bare | Slime | Encrusting Bryozoan | Calcareous Tubeworm | Colonial Tunicate | Barnacle | Sponge | Arborescent Bryozoan | Solitary Tunicate | Cnidaria |
|---|---|---|---|---|---|---|---|---|---|---|
| MAE | $3.94 \pm 0.02$ | $3.72 \pm 0.01$ | $2.58 \pm 0.01$ | $1.55 \pm 0.03$ | $2.83 \pm 0.01$ | $4.74 \pm 0.00$ | $2.17 \pm 0.01$ | $2.17 \pm 0.01$ | $4.12 \pm 0.01$ | $4.40 \pm 0.00$ |
| MAPE | $5.7 \pm 0.0$ | $48.0 \pm 0.4$ | $72.9 \pm 0.4$ | $113.6 \pm 0.5$ | $67.3 \pm 0.4$ | $20.1 \pm 0.0$ | $60.0 \pm 0.5$ | $88.1 \pm 0.5$ | $33.6 \pm 0.2$ | $24.4 \pm 0.0$ |
| LOP | $0.0 \pm 0.0$ | $17.4 \pm 0.2$ | $27.9 \pm 0.3$ | $53.7 \pm 0.3$ | $30.0 \pm 0.2$ | $0.0 \pm 0.0$ | $23.4 \pm 0.2$ | $38.4 \pm 0.3$ | $6.2 \pm 0.1$ | $0.0 \pm 0.0$ |

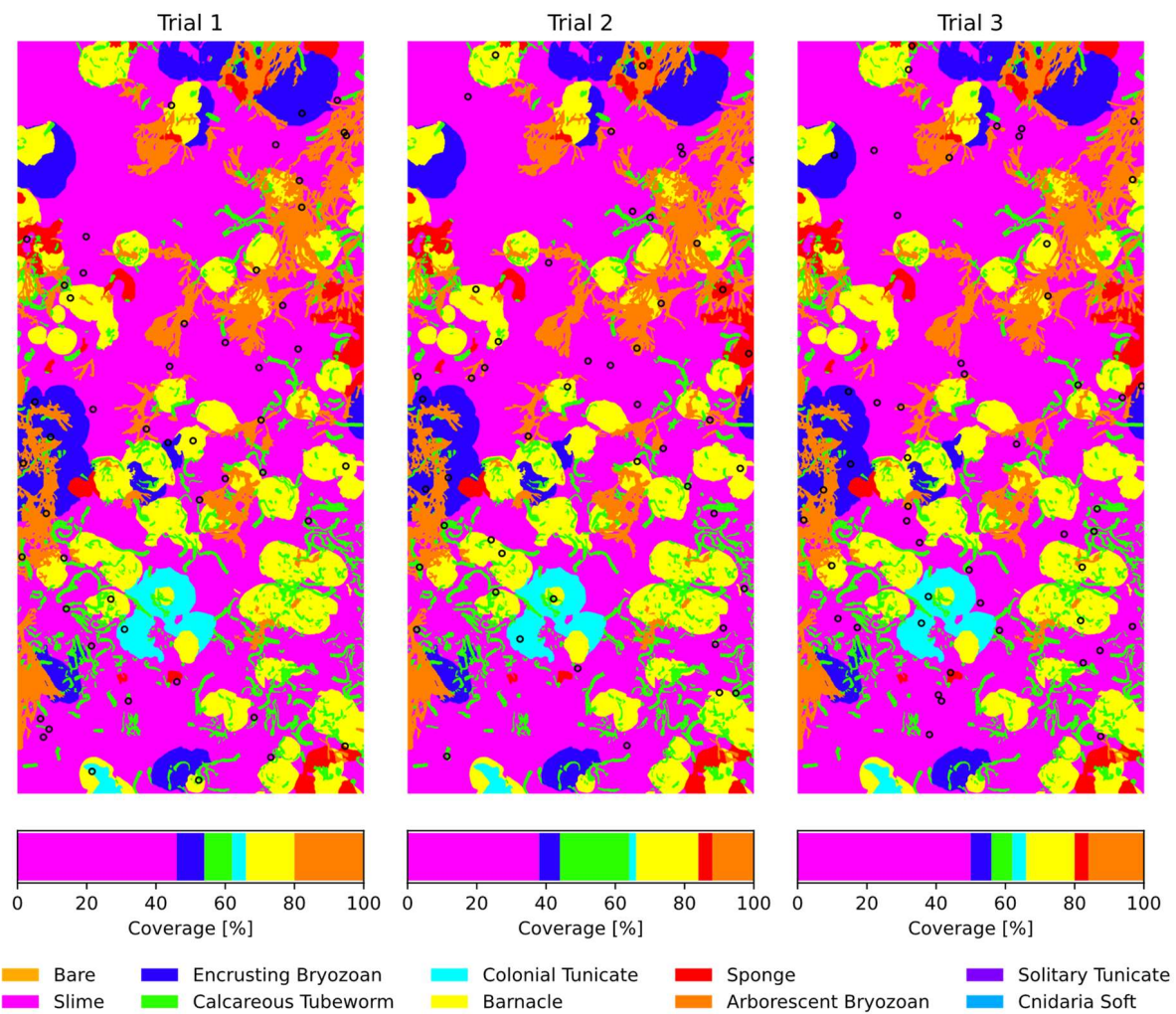

**Supplementary Figure 1: Visualization of random point sampling error.** Example of three trials of uniform random point sampling with $n = 50$ points (black circles) on the same segmentation mask acquired by manual annotation and the resulting differences in surface coverage.

**Supplementary Table 2: Segmentation performance of the enhanced U-Net.** Metrics are reported as the class-wise mean of image tiles from the validation set. Best values (≥ 85 %) are highlighted.

| Class | Bare | Slime | Encrusting Bryozoan | Calcareous Tubeworm | Colonial Tunicate | Barnacle | Sponge | Arborescent Bryozoan | Solitary Tunicate | Cnidaria | Mean |
|---|---|---|---|---|---|---|---|---|---|---|---|
| Accuracy | **0.986** | **0.911** | **0.978** | **0.981** | **0.969** | **0.976** | **0.988** | **0.982** | **0.999** | **0.993** | 0.976 |
| IoU | **0.908** | 0.807 | 0.772 | 0.644 | 0.667 | 0.666 | 0.879 | 0.505 | **0.966** | 0.672 | 0.749 |
| F1 | **0.952** | **0.893** | **0.872** | 0.784 | 0.800 | 0.800 | **0.936** | 0.671 | **0.983** | 0.804 | 0.849 |
| Precision | **0.932** | **0.885** | **0.861** | 0.805 | **0.862** | 0.801 | **0.926** | 0.709 | **0.974** | **0.858** | 0.861 |
| Recall | **0.973** | **0.901** | **0.882** | 0.764 | 0.747 | 0.799 | **0.946** | 0.637 | **0.991** | 0.756 | 0.840 |

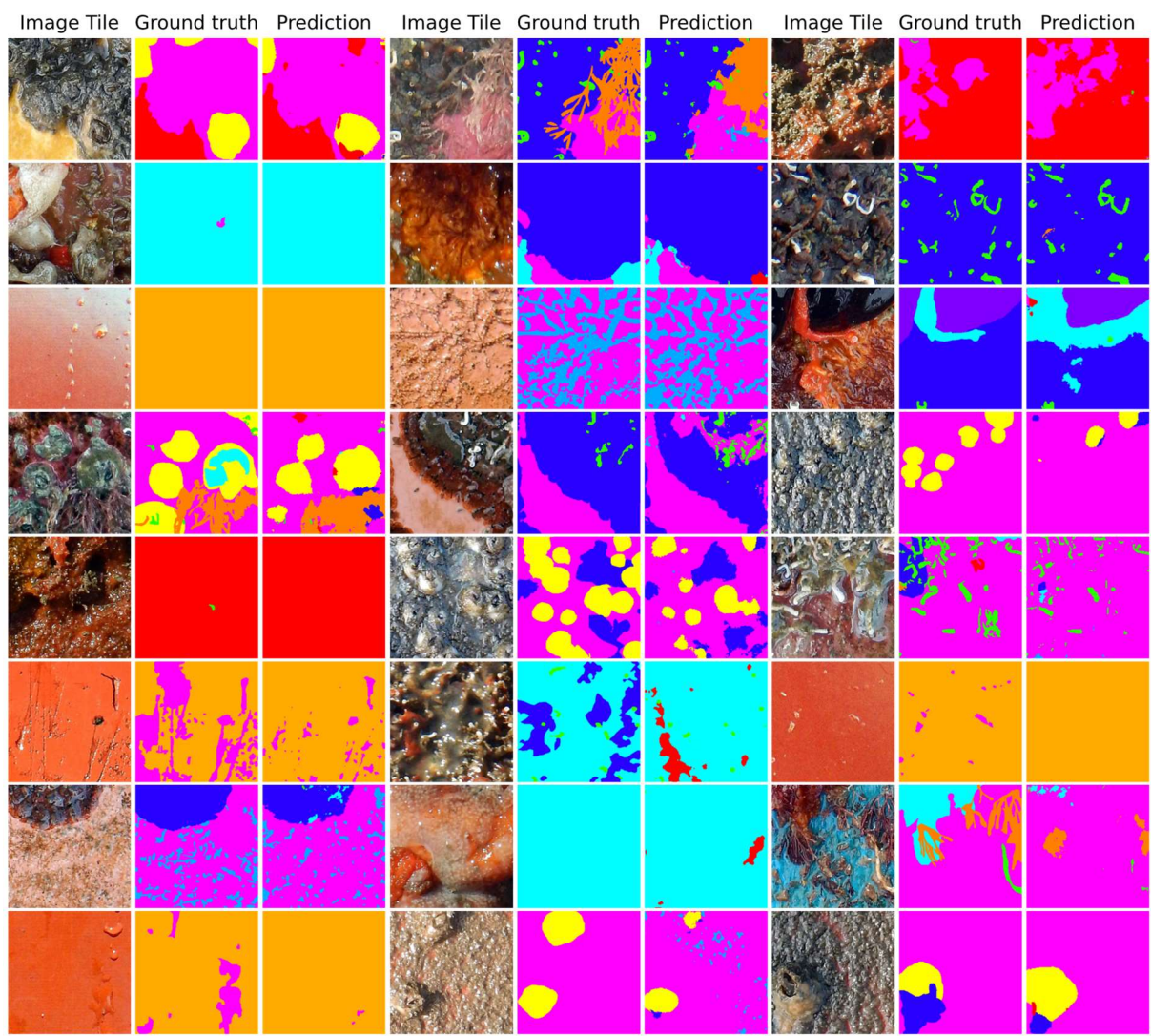

**Supplementary Figure 2: Semantic segmentation of images tiles.** Ground truth and prediction by developed model for randomly selected tiles from validation set.

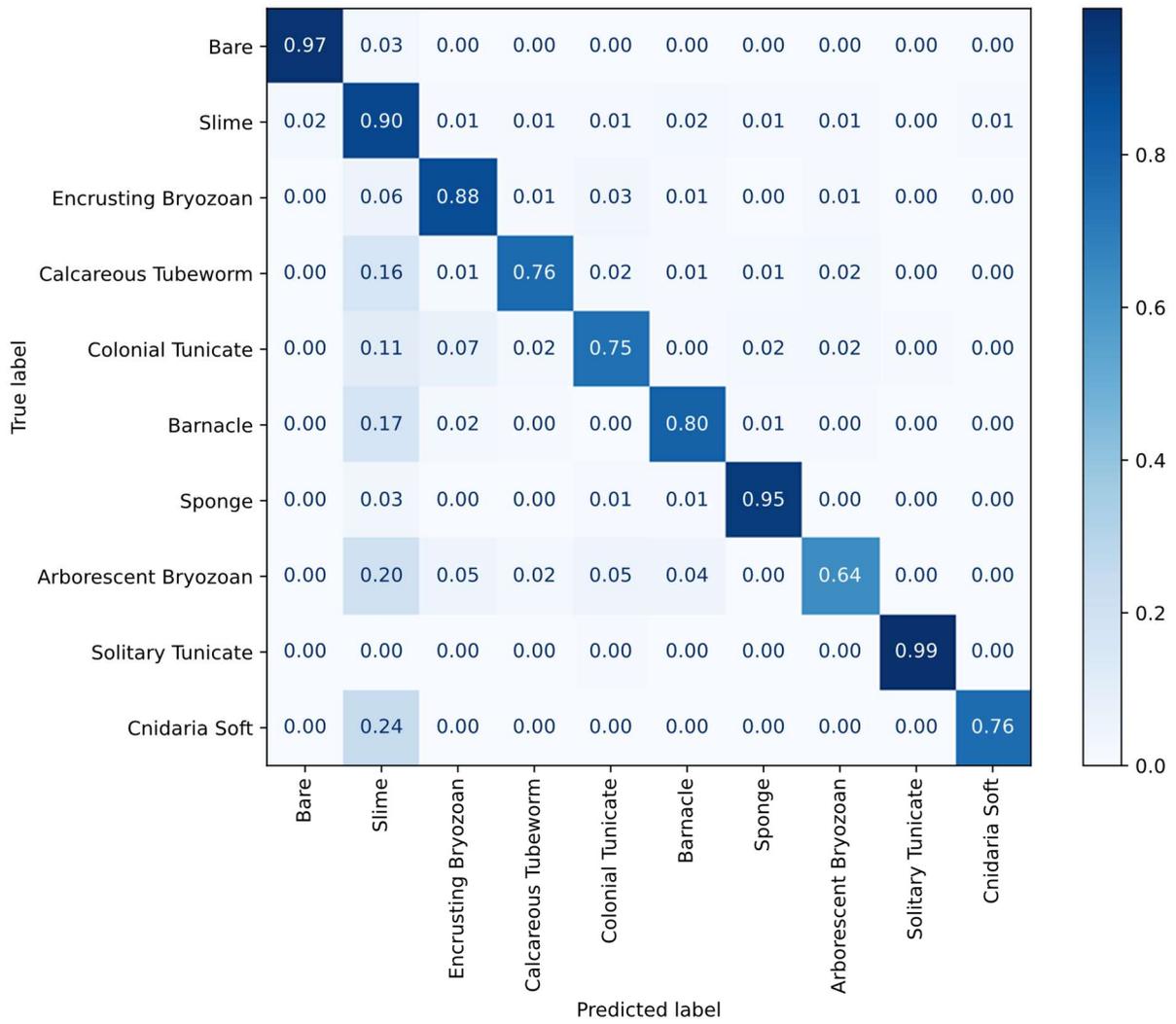

**Supplementary Figure 3: Confusion matrix of best performing model.** Rows were normalized to one for better visibility.

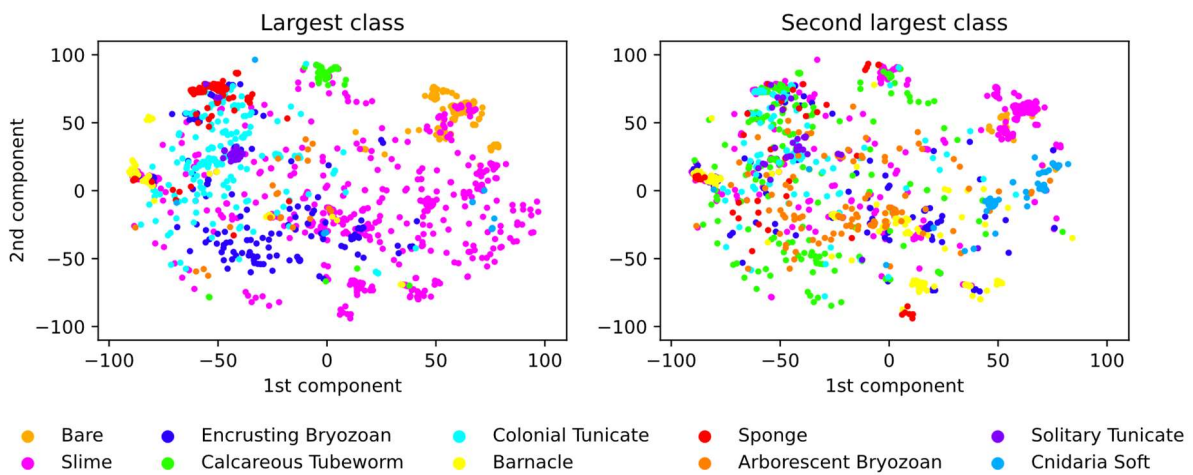

**Supplementary Figure 4: Ordering of the macrofouling classes in latent space.** For each labeled image tile, the class with the largest and second largest area was calculated. If only one class was present, the data point is missing on the right plot.

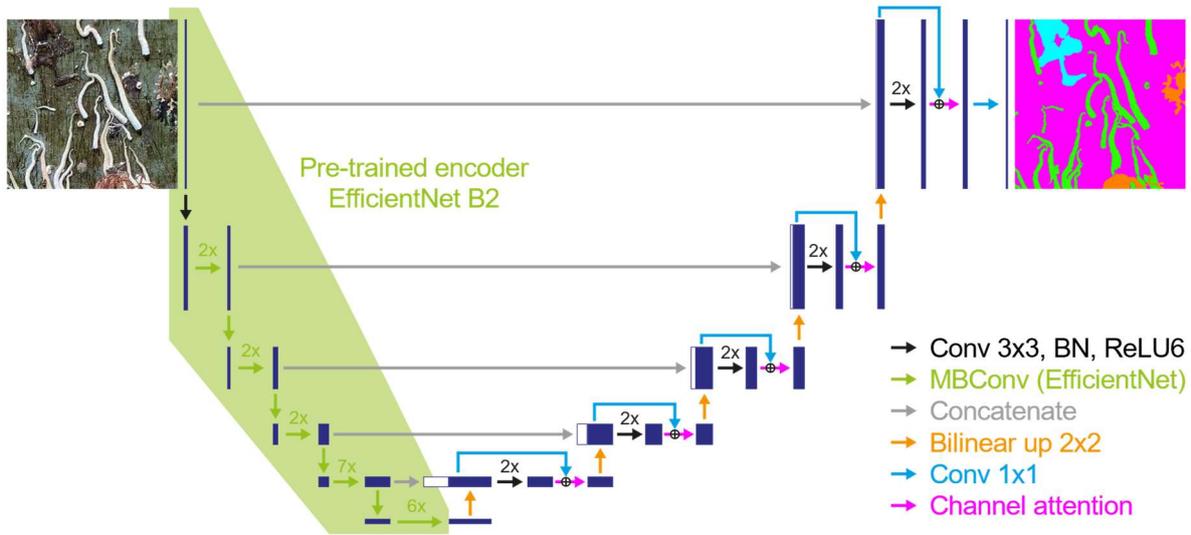

**Supplementary Figure 5: Final model architecture.** A pre-trained EfficientNet B2 encoder is deployed for the contraction path of the U-Net. Arrows denote operations and rectangles represent the resulting tensor. Crossing of arrows represents element-wise addition.